\documentclass[journal, twocolumn]{IEEEtran}
\ifCLASSINFOpdf
\else
\fi

\usepackage{float}
\usepackage[usenames,dvipsnames]{pstricks}
\usepackage[cmex10]{amsmath}
\usepackage{graphicx}
\usepackage{algorithm}
\usepackage{algorithmic}
\usepackage{amsfonts}
\usepackage{verbatim}
\usepackage{array}
\usepackage{afterpage}
\usepackage{multirow} 
\usepackage{rotating}
\usepackage[caption=false, font=footnotesize]{subfig}
\usepackage[hyphens]{url}

\begin{document}
%
\title{A Statistical Modeling Approach to Computer-Aided Quantification of Dental Biofilm}
%
%
%

\author{Awais~Mansoor,~\IEEEmembership{Member,~IEEE,}
				Valery Patsekin,
				Dale Scherl,
        J. Paul~Robinson, ~\IEEEmembership{Member,~IEEE,}
        and~Bartlomiej~Rajwa~\IEEEmembership{}
\thanks{Awais Mansoor is with the Department of Radiology and Imaging Sciences, National Institutes of Health (NIH). Prior to joining NIH, he was with the Department
of Electrical and Computer Engineering, Purdue University, West Lafayette,
IN 47906 USA. E-mail: awais.mansoor@gmail.com.}
\thanks{Valery Patsekin, and Bartlomiej Rajwa are with Bindley Bioscience Center, Purdue University, West Lafayette, IN 47906 USA. Tel: +1 765 494-0757, Fax: +1 765 494-0517, E-mail: \{brajwa\}@purdue.edu}\thanks{J. Paul~Robinson is with Weldon School of Biomedical Engineering, Purdue University, West Lafayette, IN 47906 USA. Tel: +1 765 494-0757, Fax: +1 765 494-0517. E-mail: wombat@purdue.edu.}
\thanks{Dale Scherl is with Hill's Pet Nutrition. 	Topeka, KS, USA.}
}

%
%

\markboth{}%
{Shell \MakeLowercase{\textit{et al.}}: Bare Demo of IEEEtran.cls for Journals}
%



\maketitle

\begin{abstract}
Biofilm is a formation of microbial material on tooth substrata. Several methods to quantify dental biofilm coverage have recently been reported in the literature, but at best they provide a semi-automated approach to quantification with significant input from a human grader that comes with the grader's bias of what is foreground, background, biofilm, and tooth. Additionally, human assessment indices limit the resolution of the quantification scale; most commercial scales use five levels of quantification for biofilm coverage (0\%, 25\%, 50\%, 75\%, and 100\%). On the other hand, current state-of-the-art techniques in automatic plaque quantification fail to make their way into practical applications owing to their inability to incorporate human input to handle misclassifications. This paper proposes a new interactive method for biofilm quantification in Quantitative light-induced fluorescence (QLF) images of canine teeth that is independent of the perceptual bias of the grader. The method partitions a QLF image into segments of uniform texture and intensity called \emph{superpixels}; every \emph{superpixel} is statistically modeled as a realization of a single 2D Gaussian Markov random field (GMRF) whose parameters are estimated; the superpixel is then assigned to one of three classes (\emph{background, biofilm, tooth substratum}) based on the training set of data. The quantification results show a high degree of consistency and precision. At the same time, the proposed method gives pathologists full control to post-process the automatic quantification by flipping misclassified superpixels to a different state (background, tooth, biofilm) with a single click, providing greater usability than simply marking the boundaries of biofilm and tooth as done by current state-of-the-art methods.
 
\end{abstract}

\begin{IEEEkeywords}
Biofilm, dental plaque, quantitative fluorescence, superpixelization, Gaussian Markov random field. 
\end{IEEEkeywords}

%
\IEEEpeerreviewmaketitle

\section{Introduction}
\IEEEPARstart{T}{he} assessment of dental biofilm and dental plaque coverage is a crucial step in many clinical as well as basic biological science applications.  For instance, clinicians diagnosing potential gingival conditions or evaluating dental prostheses or dental implants require an accurate estimation of bacterial biofilm coverage. Furthermore, researchers involved in studies of periodontal diseases and periodontal therapies as well as those assessing oral hygiene products often employ information about biofilm depth, area, and distribution. 

Dental-plaque images have traditionally been collected using reflected light. The quantitative evaluation is then performed manually by outlining the plaque areas using tracing paper, and subsequently quantifying the results by assigning a dental score \cite{Ramfjord195951}. In the past, a number of manual dental-plaque scoring indices have been proposed \cite{Kang2010470, Kang200778, Hennet2006175, Smith20011158, Sagel2000130}. For instance, the plaque index used by Quigley and Hein, and modified by Turesky, is one of the most frequently employed to grade images. Others in general use are the Ramfjord index, the Turesky index, and the O'Leary index \cite{Pretty2005193}.

During the past decade, computer assisted technology have gained a lot of interest in almost every aspect of biology and medicine from accessibility \cite{mansoor2010accessscope} to quantitative measurements \cite{foster2014review, xu2013559, sandouk2013accurate} to qualitative adjustments \cite{mansoor2014noise}. With the introduction of digital imaging, the idea of computer-aided quantification in dentistry and dental research began to gain popularity. However, early computer-aided techniques presented in the literature failed to offer practical benefits to researchers, as almost all reported computer-aided quantification techniques relied on manual segmentation performed by a grader using a software tool. These methods employing commercial photo-editing software are tedious, are time consuming, and do not provide a tangible advantage over manual grading. For instance, in \cite{Kang200778, Hennet2006175} an approximate region of interest (ROI) in a tooth imaged under reflected light is manually selected using Photoshop (Adobe, San Jose, CA) tools; a mean-shift segmentation algorithm is then employed to estimate the percentage of tooth surface covered by dental plaque. Smith et al. \cite{Smith20011158} use a \emph{pen} and \emph{make path} tools of Photoshop. Subsequently ImagePro Plus (Media Cybernetics, Silver Spring, MD) was used for calibration and calculation of the percentage of total tooth area covered by plaque.

Partly owing to the low quality and low contrast of reflected-light images, the early attempts at computer-aided dental diagnostics did not prove successful. A major breakthrough in automated dental-plaque quantification methods occurred with the introduction to dentistry of fluorescence dyes that stain the bacterial biofilm, providing contrast between the region of interest and background. UV-excited fluorescein produces a yellow plaque signal, which contrasts with the blue tooth enamel signal and with the gingiva, which appears black in images. This makes fluorescein an excellent candidate for automatic quantification of biofilm, and methods based on fluorescein imaging have been proposed in the literature. For instance, Sagel et al. \cite{Sagel2000130} employ UV fluorescence imaging and automatically classify every pixel inside the image by calculating the scalar distance in RGB color space from the pixel to the median RGB color of each predefined class  \cite{Sagel2000130}. The pixel is then assigned to the closest group. 

\begin{figure}[htb]
\centering
\includegraphics[scale=0.7]{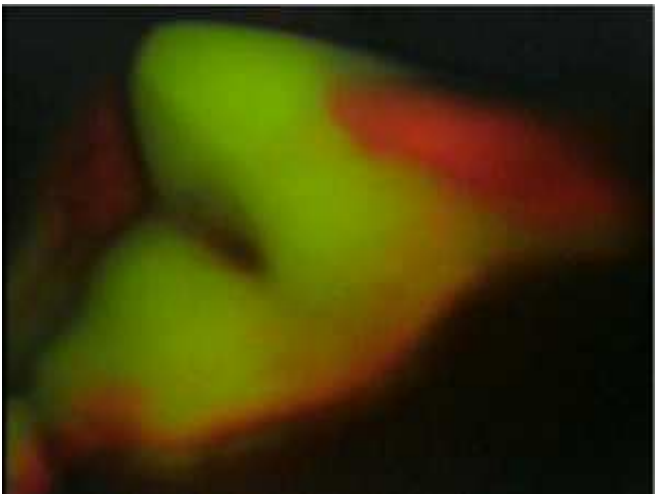}\\
\caption{Disclosed plaque imaged using QLF instrument.}
\label{fig:fluorescein}
\end{figure}

Quantitative light-induced fluorescence (QLF) is yet another dental diagnostic tool for quantitative assessment of caries, plaque, bacterial activity, and staining \cite{pretty2002105}. The QLF methods are based on the autofluorescence of teeth and plaque-forming bacterial biofilm. The QLF devices marketed by Inspektor Dental Care BV (Amsterdam, The Netherlands) have recently generated considerable interest in the dental research community \cite{amaechi20027}. The technique uses a small camera that can be easily hand held; the images are free from flash-light specular reflections and major distortions. Additionally, the QLF technique enables very small changes in plaque to be detected, thus increasing accuracy \cite{Pretty2005193}. Going beyond mere detection, QLF can monitor the development or regression of caries and plaque over time. When teeth are excited with blue light, the enamel emits green luminescence, the dental plaque red. This red/orange fluorescence is attributed to metabolic products (mostly porphyrins) from some of the resident bacteria. The intensity of autofluorescence is expected to be proportional to the biofilm depth (Fig. \ref{fig:fluorescein}). A recent study by Pretty et. al \cite{pretty2012336} confirmed the potential of digital imaging systems employing polarized white light and QLF for software analysis and assessment by raters for  epidemiological work. Owing to these, QLF imaging has of late become a popular modality for plaque quantification, and a number of QLF image--based quantification methods have been proposed \cite{Pretty2002158, Tranæus200171, Pretty20031151}. 

With the advent of fluorescence-imaging techniques in dentistry several attempts have been made to automate the quantification process, and a number of statistical and machine-learning approaches have been tested for the purpose of automation in an attempt to minimize the subjectivity of manual scoring. For instance, \cite{Kang2006797} used a cellular neural network to search for the threshold between biofilm and clean areas of the tooth; \cite{Kang200778, Kang2010788, Jiayin20103068} quantified plaque by using \emph{fuzzy c-means} clustering. However, the implementation of functional and robust automated segmentation in the context of dental images turned out to be an unexpectedly difficult problem. The demarcation of a tooth of interest from neighboring teeth fluorescing almost identically, as well as from gum tissue, still needs to be investigated further, and current state-of-the-art mathematical models fail to accurately mimic human expert assessments. The problem is significantly worse in veterinary dental imaging owing to the lower image quality caused by difficulties associated with animal compliance. 

Ambitious attempts to replace image graders or qualified pathologists with computer systems usually fail owing to lack of algorithm robustness, but also because of tradition and praxis. On the other hand, digital image-analysis platforms designed for general use do not provide practitioners with tools that rely on and supplement their expertise. Consequently, current state-of-the-art automatic plaque-quantification approaches have not gained much popularity owing to the fact that they allow only the choice between accepting or rejecting the automated quantification; even in cases where the result is off by only a little, the user typically has no option of correcting the misclassification but must reject it altogether and resort to a painstaking method of manual demarcation.

The method outlined in this paper employs a combination of manual and automated segmentation, providing what can be considered essentially a computer-aided segmentation and quantification. In contrast to a completely automated quantification approach, the described method provides an initial estimation of the background, the teeth, and the plaque-covered area based on the prior model of expert classification embedded into the method. This first approximation that happens to be very accurate for most cases, can be subsequently modified by a grader or dental research practitioner in an interactive manner if deemed necessary. Additionally, the interaction does not require tedious manual re-segmentation. Instead, the operator points to entire computer-suggested regions to assign them to one of the pre-defined categories (biofilm, tooth subtrata, background). This approach allows for much faster and more streamlined interaction between operator and computer system. The availability of the initial assessments provided by the algorithm allows for rapid quantification with a minimal number of significant corrections. The described technique has been practically tested and validated for quantification of dental film on canine teeth.

The approach presented in this paper, which combines automated and interactive functionality, represents a broader solution to image-analysis problems in clinical and research pathology, and is especially applicable to challenging, difficult-to-segment biomedical images. The described technique is also amenable to further extension employing machine-learning methods. 

\begin{figure}[htb]
\centering
\includegraphics[scale=0.3]{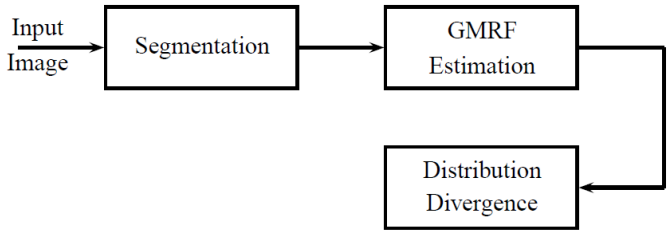}
\caption{A block diagram summarizing the proposed algorithm for quantification of dental biofilm on canine teeth.}
\label{fig:blockdiagram1}
\end{figure}

The basic principle of the utilized algorithm relies on its ability to divide images into statistically and texturally uniform regions called superpixels, which can be subsequently compared using a measure of dissimilarity \cite{mansoor2012}. The block diagram of the algorithm is shown in Fig. \ref{fig:blockdiagram1}. The method works in two stages: (i) \emph{initial segmentation}, and (ii) \emph{interactive correction}. During the initial segmentation stage the superpixelization process creates a set of statistically uniform regions, which are classified into three categories (background, clean area (tooth substratum), and biofilm-covered area) using statistical texture estimation based on GMRF. The initial guess, imitating the expert quantification, is based on a calibration that employs a number of manually scored clean and covered areas from various teeth. Owing to the calibration this initial segmentation process offers reasonably high concordance with manual grading. However, in the interactive corrections stage, the segmentation can be further modified by operators by pointing to a questionable region and clicking the mouse. This methodology offers much higher efficiency than a requirement for manual drawing of outlines or a requirement to redraw contested areas. Moreover, as shown later in the results section, the correction needs to be applied only to one image in a series, the subsequent images may be re-segmented automatically using the corrected mask. 

\section{Background}
As mentioned in the introduction, general-use photo-editing packages fail to provide intuitive assistance to biomedical image analysts such as pathologists or dental image graders. The published procedures often need a precise outline to be drawn around tooth and/or biofilm. However, with the advancement of human-computer interaction techniques, it becomes clear that creating a link between human grader experience and automated system should lead to improvement in computer-aided dental diagnostics. The method proposed here presents a tool that quantifies biofilm using an interactive segmentation interface that allows practitioners to quickly refine the initial assessment suggested by the algorithm.

The principle of the proposed automatic quantification method is based on the fact that a uniform region inside an image can be modeled via a Gaussian Markov random field (GMRF), a statistical image-modeling technique that has long been used for statistical image segmentation \cite{Krishnamachari1995} and classification \cite{Chellappa1985959}. The statistical modeling serves two purposes in terms of dental biofilm quantification: it will help isolate the tooth inside the image and it will segment the plaque region inside the tooth. The first step is to segment the image into sub-regions uniform enough that they can be modeled using a single GMRF; therefore we start with a brief introduction to the concepts of the oversegmentation method proposed by Malik et al. \cite{Jianbo2000888, Rue1984}.

\subsection{Oversegmentation (Superpixelization)}
\label{subsec:superpixel}
The idea of oversegmentation by dividing an image into what is known in the literature as \emph{superpixels} was proposed by Ren and Malik \cite{Ren200310}. Their motivation was to come up with a unit of \emph{visual} representation having uniform textural and statistical properties, since digital pixels are units of \emph{digital} representation. Since the inception, the idea has been used in various applications of image processing \cite{Barnard20031107, Felzenszwalb2004167, Deng2001800}, machine learning \cite{Blum2001, Weinberger2009207}, and computer vision \cite{Stauffer2000747}. Our algorithm uses this method as a preprocessing step to GMRF estimation. The superpixels are texturally and statistically more meaningful than regular pixels in that each superpixel forms a perceptually consistent unit, i.e., all pixels inside a superpixel are assumed to be consistent in their statistical properties and texture  \cite{Jianbo2000888, Felzenszwalb2004167, Shi1998943}. In our method, the superpixel segmentation map not only helps segregate biofilm from the tooth subtrata, but also eliminates the need for any independent background subtraction procedure.

\begin{figure}[htb]
\centering
\subfloat[][]{\includegraphics[scale=0.5]{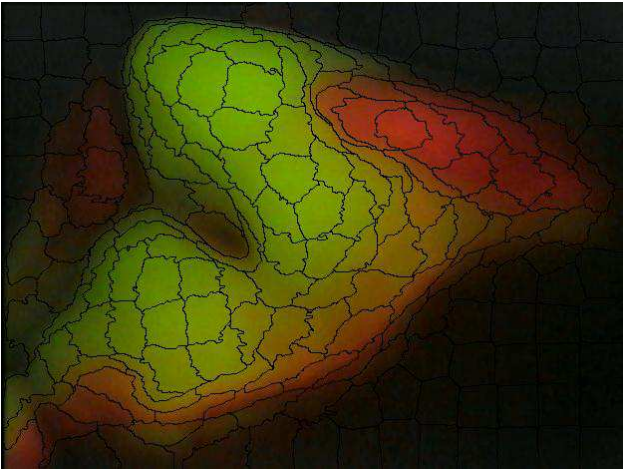}\label{fig:superpixel200}}\\
\subfloat[][]{\includegraphics[scale =0.7]{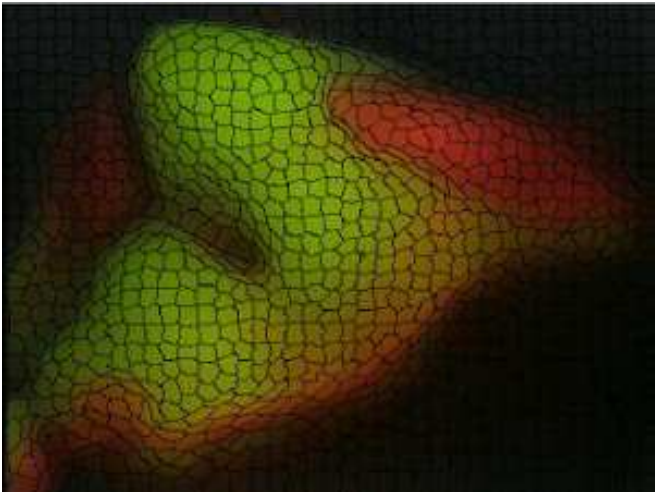}\label{fig:superpixel1000}}
\caption{An example of superpixels obtained through the \emph{normalized cut} algorithm \cite{Jianbo2000888}. (a) is segmented into $200$ superpixels; (b) is segmented into $1000$ superpixels.}
\label{fig:superpixels}
\end{figure}

The \emph{graph cut} method for superpixelization \cite{Jianbo2000888, Shi1998943} uses the graph-theory concept of iterative binary partitioning: each pixel is modeled as a vertex of the graph, and a weighted edge between two pixels shows the degree of similarity (or dissimilarity) between them. Within this model, the image-segmentation problem is reduced to partitioning a connected graph into several disjoint subgraphs called \emph{graph cuts}, and the nodes of the graphs are the entries that are partitioned. In the case of images, the pixels form the nodes of the graph, and the edges between neighboring nodes correspond to the \emph{strength} with which two pixels belong to the same segment. The criterion for partitioning is to minimize the weights of the connections within each group and maximize the weights among different groups. The input arguments required by the algorithm to produce an image segmentation map are the image and the number of groups (\emph{superpixels}). Fig. \ref{fig:superpixels} shows images segmented into 200 and 1000 superpixels. The typical number of superpixels per image depends upon the size of the image and the textural details inside the image and determined empirically by the user, we have not found any evidence of the dependence of the number of superpixels on patients in our experiments. The selection to the number of superpixels need to be made once only for a typical dataset. If the superpixels are too small, the second step of manually relabeling them takes a lot of time, on the contrary if they are too large the precision of the estimated parameters of the random field in the next stage is compromised. Empirically in our experiments and as shown later in Fig. \ref{fig:superpixels}, for QLF images of size $256\times 256$, partition to 200 superpixels is found sufficient for dental quantification purposes. 

\subsection{GMRF Modeling of QLF Images}
After obtaining superpixels for an image, the proposed method models every superpixel as a realization of GMRF. GMRF model is used to capture the statistical properties of the neighborhood around a pixel. A 2D Gaussian random field (GRF) is a random vector following a multivariate normal distribution; putting an additional constraint of conditional independence makes it \emph{Markovian}, which essentially means that the statistics of a pixel inside an image are independent of those of the pixels outside a predefined neighborhood, hence the term Gauss Markov random field. Mathematically, for two indices $i$ and $j$ inside the images,
\[
x_i  \bot x_j |\mathbf{x}_{ - ij} 
\]
where $x_i$ and $x_j$ are conditionally independent of each other given the neighborhood $\mathbf{x}_{ - ij}$, ($\mathbf{x}_{ - ij}$ denotes all the elements inside the neighborhood but not including $i$ and $j$). Let $\mathcal{G}=(\mathcal{V},\mathcal{E})$ be an undirected graph where $\mathcal{V}={1,2,\dots}$ are vertices of the graph and $\mathcal{E}=({i,j})$ are the edge pair of the graph; then a random vector $\mathbf{x}=(x_1,x_2,\dots,x_n)^T$ is called the GMRF with respect to the graph $\mathcal{G}$ with 
mean $\mathbf{\mu}$ and covariance matrix $\Sigma$ if its density has the form
\[
\Pi \left( \mathbf{x} \right) = \left( {2\pi } \right)^{ - {n \mathord{\left/
 {\vphantom {n 2}} \right.
 \kern-\nulldelimiterspace} 2}} \left| \Sigma  \right|^{ - {1 \mathord{\left/
 {\vphantom {1 2}} \right.
 \kern-\nulldelimiterspace} 2}} \exp \left( { - \frac{1}{2}\left( {\mathbf{x} - \mathbf{\mu} } \right)^T \Sigma^{-1} \left( {\mathbf{x} - \mathbf{\mu} } \right)} \right)
\]

\subsection{Parameter Estimation of GMRF}
The modeling of 2D homogeneous texture is connected to a 2D symmetric autoregressive process driven by a white noise process $E_\mathbf{s}$, $\mathbf{s} \in \mathbb{R}^2$ \cite{Descombes1999490, Ranguelova1999430}. For any random field, a 2D symmetric autoregressive process driven by a white noise can be written as 
\begin{equation}
A_\mathbf{s}  = B_\mathbf{s} \theta _\mathbf{s}  + E_\mathbf{s} 
\label{eq:s3e1}
\end{equation}
where $\theta_\mathbf{s}$ are symmetric autoregressive parameters calculated over a predefined neighborhood, and $A_\mathbf{s}$ is a discrete-valued random field defined on a 2D lattice. $B_\mathbf{s}$ is the neighborhood matrix such that every element  $b_\mathbf{s}  = \left[ {a_{\mathbf{s} + \mathbf{r}}  + a_{\mathbf{s} - \mathbf{r} ,\mathbf{r}}} \right]$, $\eta$ is a predefined neighborhood of radius $\mathbf{r}$. Owing to the Markovian assumption and the superpixel segmentation the parameters of the random field are independent of the pixels outside the predefined neighborhood $\eta$. To estimate the parameters of the Gauss-Markov model we used a least-squares (LS) approach on a predefined neighborhood $A_\mathbf{s}, \mathbf{s}\in \eta$. The least-squares minimization to (\ref{eq:s3e1}) with respect to $\theta_\mathbf{s}$ is
\begin{eqnarray} 
\theta _\mathbf{s}  =  {\mathop {\arg \min }\limits_{\theta _\mathbf{s} }} E_\mathbf{s}^T E_\mathbf{s} = \left( {B_\mathbf{s}^T B_\mathbf{s} } \right)^{ - 1} B_\mathbf{s}^T A'_\mathbf{s} 
\end{eqnarray}

where $A'_\mathbf{s}=A_\mathbf{s}-\hat \mu_\mathbf{s}$ and $\mathop {\hat \mu_\mathbf{s}  = \left( {N\times M} \right)}^{ - 1} \sum\nolimits_{\mathbf{s} \in \omega _\mathbf{r} } {a_\mathbf{r} } $, $N\times M$ are the dimensions of the neighborhood $\eta$ (one can use arbitrary dimensions but, in practice, $M=N$). The variance estimate $\hat \sigma_\mathbf{s}^2$ can be calculated in a similar manner:

\begin{eqnarray}
\hat \sigma_\mathbf{s}^2  = \frac{1}{{NM}}\sum\limits_{\mathbf{s} \in \eta _\mathbf{r} } {\left( {a_\mathbf{s}  - \mathbf{b}_\mathbf{s}^T \hat \theta _\mathbf{s} } \right)^2 } \nonumber\\
\hat \Sigma _\mathbf{s}  = \frac{1}{{NM}}\left( {A_\mathbf{s}  - B_\mathbf{s} \hat \theta_\mathbf{s} } \right)^T \left( {A_\mathbf{s}  - B_\mathbf{s} \hat \theta_\mathbf{s} } \right)
\end{eqnarray}

An inscribed bounding-box is used inside every superpixel for texture estimation. The least-squares estimate provides a feature vector $f_\mathbf{s} = \left[ { \mathbf{\theta}_\mathbf{s} , \hat \Sigma_\mathbf{s}  ,\mathbf {\hat \mu}_\mathbf{s}} \right]$. It is important to note here that the method assumes homogeneity of texture inside a predefined neighborhood, achieved through superpixelization.

\subsection{KL Divergence}
After estimating the parameters of the GMRF, we use Kullback-Leibler (KL) divergence to express the degree of similarity between two probability distributions. Let $\mathcal{C}_{1\Lambda}(n)$ and $\mathcal{C}_{2\Lambda}(n)$ be the sets $2D$ random-field estimates inside two channels $\mathcal{C}_1$ and $\mathcal{C}_2$ respectively; then the KL divergence between them is defined as
\begin{equation}
\label{eq:KLD}
KL\left( {\mathcal{C}_{1\Lambda}||\mathcal{C}_{2\Lambda}} \right) = \sum\limits_{n = 1}^\mathcal{N} {\mathcal{C}_{1\Lambda}\left( n \right)\log \frac{{\mathcal{C}_{1\Lambda}\left( n \right)}}{{\mathcal{C}_{2\Lambda}\left( n \right)}}}
\end{equation}
where $n\in\mathcal{N}$ is the superpixel index, $\mathcal{N}$ is the total number of superpixels.

Since KL divergence is not symmetric, we make Eq. (\ref{eq:KLD}) symmetric by calculating both $KL\left( {\mathcal{C}_{1\Lambda}||\mathcal{C}_{2\Lambda}} \right)$ and $KL \left( {\mathcal{C}_{2\Lambda}||\mathcal{C}_{1\Lambda}} \right)$. The symmetric variant of the $KL_{S}$ is defined as
\begin{equation}
\label{eq:symKL}
KL_{S}\left( {\mathcal{C}_{1\Lambda},\mathcal{C}_{2\Lambda}} \right)=KL\left( {\mathcal{C}_{1\Lambda}||\mathcal{C}_{2\Lambda}} \right) + KL \left( {\mathcal{C}_{2\Lambda}||\mathcal{C}_{1\Lambda}} \right)
\end{equation}
From \cite{Muirhead1982}, the KL divergence for multivariate normal densities $\mathcal{C}_{1\Lambda}\left( {\mu_{\mathcal{C}1} ,\Sigma_{\mathcal{C}1} } \right)$ and $\mathcal{C}_{2\Lambda}\left( {\mu_{\mathcal{C}2} ,\Sigma_{\mathcal{C}2}} \right)$ forming a Markov random field (i.e. GMRF) is
\begin{eqnarray}
KL\left( {\mathcal{C}_{1\Lambda}||\mathcal{C}_{2\Lambda}} \right)&=&0.5\log \frac{{\left| {\Sigma_{\mathcal{C}1} } \right|}}{{\left| {\Sigma_{\mathcal{C}2} } \right|}} + 0.5Tr\left( {\Sigma_{\mathcal{C}1}^{ - 1} \Sigma_{\mathcal{C}2} } \right)\\&+&0.5\left( {\mu_{\mathcal{C}1}  - \mu_{\mathcal{C}2} } \right)^T \Sigma_{\mathcal{C}1}^{ - 1} \left( {\mu_{\mathcal{C}2}  - \mu_{\mathcal{C}1}}  \right) - \frac{d}{2}\nonumber
\label{eq:NormKL1}
\end{eqnarray}
where the operator ${\left| . \right|}$ denotes the determinant of a matrix and $d$ is the dimension of the covariance matrix. 

\section{Automatic Quantification Algorithm for Biofilm Coverage}
\begin{figure*}[!htb]
\vspace{0.5in}
\centering
\subfloat[Calibration stage]{
\includegraphics[scale=0.4]{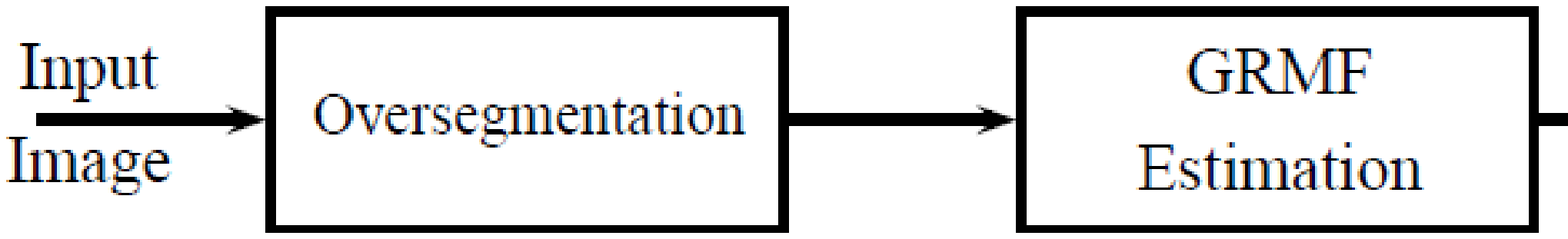}
}

\subfloat[Quantification stage]{
\includegraphics[scale=0.35]{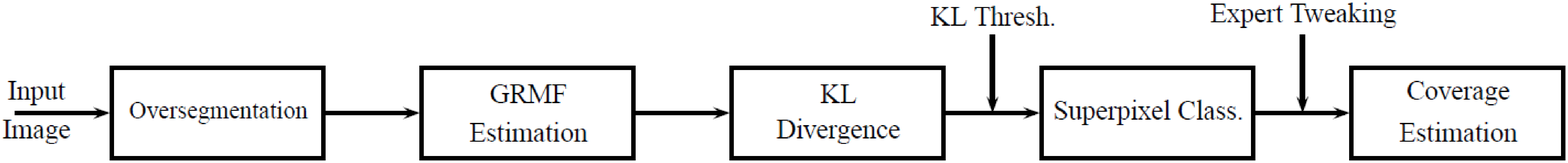}
}

\caption{A block diagram summarizing the proposed algorithm for quantification of dental biofilm. The diagram shows both stages of the proposed algorithm: (a) \emph{the calibration stage} and (b) \emph{the quantification stage}.}
\label{fig:blockdiagram2}
\end{figure*}

In the previous section we presented the theoretical basis for our algorithm for quantification of biofilm on canine teeth. This algorithm is summarized in the block diagram shown in Fig. \ref{fig:blockdiagram2}. The algorithm has two stages: the \emph{calibration} stage and the \emph{quantification} stage. The proposed method begins by converting the RGB image into HSI (\emph{hue, saturation, intensity}) domain and computing superpixels (oversegmentation) inside the \emph{intensity} channel of a QLF image. The superpixel map obtained is mapped over the green channel, followed by the estimation of statistical parameters of every superpixel individually in both channels. The idea is to assume every superpixel to be a realization of a unique GMRF, estimate the parameters of the random field in two channels, and compare their statistics using a divergence measure. As explained before, in QLF images the biofilm areas of the tooth are captured primarily in the red channel and the clean areas (tooth substratum) in green. However, the tooth substratum masked by biofilm does not register inside the green channel; therefore, we used the intensity channel and the green channel for random-field estimation and divergence estimation.  The task is the isolation of three classes inside the image: biofilm, tooth substratum, and background. This is challenging owing to the presence of multiple neighboring teeth in the background, ambient illumination, and  imaging artifacts near the gums. Most biofilm quantification methods proposed in the literature require background subtraction as a preprocessing step. However, our method does not need any independent background subtraction technique to isolate a tooth from the image, owing to the statistical modeling approach presented in this paper that can separate background, tooth substrata, and the biofilm based solely on the parametric differences of their estimated Markov random field model. The algorithm takes a tree-like approach to classify the image areas as \emph{biofilm}, \emph{tooth substratum}, and \emph{background} by first separating the \emph{background} from the target dental field, and then segmenting the \emph{tooth substratum} from the \emph{biofilm}. The extent of divergence between superpixels from the green channel and the total intensity channel will determine if a particular superpixel belongs to biofilm or not. The absence of biofilm corresponds to both channels being identical in terms of intensity and texture, and therefore to a low statistical divergence value. In the presence of biofilm, superpixels will produce a high statistical divergence. The KL divergence values for the background and the clean tooth substratum will be close, so we added another feature for classification: the mean intensity of the superpixel. After classifying biofilm superpixels, the algorithm performs binary classification for the rest of the superpixels based on two features: KL divergence and mean illumination intensity. Fig. \ref{failedimages} shows a single tooth isolated by superpixelization and the tooth-isolation procedure of our algorithm. To account for multiple teeth inside an image, it is further assumed that the tooth of interest lies close to the geometric center of the image.

\subsection{Calibration Stage}
Expert opinions are usually used as the ground-truth in medical and biological imaging, the calibration stage in our method is a way of incorporating the expert-knowledge into the system. Once the expert determines the boundary between the biofilm and the tooth substrata; our method by extracting the statistical differences of the two class, imitates the human expert for images acquired in similar settings. The block diagram in Fig. \ref{fig:blockdiagram2} outlines the calibration procedure. The aim of the calibration stage is to establish the KL-divergence threshold between tooth substratum and biofilm. To define clean areas we utilize images which are deemed clean by two expert graders. The experts used in our experiments for manual scoring possess several years of industrial experience in acquisition, calibration, and evaluation of QLF images. 

After the \emph{tooth-isolation} step that separates tooth from background based on two features, i.e. KL-divergence, and the mean intensity of superpixel, the calibration sets the KL threshold equal to the $95^{th}$ percentile of KL divergences among superpixels inside the clean teeth. The percentile is chosen empirically to accommodate human bias and hardware and capturing related issues such as inappropriate lighting. Since the KL divergences are calculated between the green channel and the intensity channel, the KL-divergence values increase with the increase in the red hue inside the superpixels. It is up to the human expert to decide how low the red signal level must be in order to declare the substratum region clean. The ground truth from the expert segmentations are created using the STAPLE algorithm \cite{warfield2002298}. 

\subsection{Quantification Stage}
The \emph{quantification stage} follows the same steps as the \emph{calibration stage}. After the \emph{tooth-isolation} procedure, the quantification stage compares the KL divergence of isolated-tooth superpixels with the KL divergence threshold of tooth substratum and plaque calculated in the calibration stage. The \emph{biofilm quantification index} (BQI) defines the percent area of tooth covered with plaque; mathematically
\begin{equation}
BQI = \frac{{\sum\limits_{t \in plaque} {\left| {I_{sp} \left( t \right)} \right|} }}{{\sum\limits_{t \in tooth} {\left| {I_{sp} \left( t \right)} \right|} }}
\end{equation}
where the $\left| . \right|$ operator gives the area of an individual superpixel.

\section{Imaging Platform}
Images of canine teeth were captured using a quantitative light-induced fluorescence (QLF) system from Inspektor Research Systems.  This system is equipped with a Philips MPXL RP50-2P xenon arc lamp, Hoya HA30 infrared band-pass filter, Hoya B370 blue bandwidth filter, Lumatec S380 liquid light guide, Hoya Y52 yellow high-pass filter, Sony DXC-LS1P CCD camera system, Integral Technologies FlashPoint 3DX Lite PCI framegrabber, and Inspektor Research Systems software version 2.0.38.  Image capture was perform on conscious dogs, with the room lights dimmed in order to reduce interference from visible light.  The camera uses a fiber-optic cable in combination with a light shield and mirror that allows it to be positioned in the mouth with an orientation that best facilitates image capture of each tooth. QLF imaging uses a high-intensity blue light (405 $\pm$ 20 nm) that is shined onto the tooth surface and causes clean, undamaged enamel to fluoresce green.  Red/orange fluorescence can also be seen on the tooth surface from within the oral biofilm.

\section{Results and Discussion}
To evaluate the performance of our approach, we tested the quantification of dental plaque on canine teeth using QLF images.  In order to evaluate the quality of the initial starting segmentation provided by the algorithm, another test is created where the automatic quantification results were compared to the rater scoring at 8 grades. The human experts assigned teeth imaged using QLF to 8 classes based on the amount of plaque covering the teeth ($0\%$, $12.5\%$, $25\%$, $37.5\%$, $50\%$, $62.5\%$, $75\%$, and $>87.5\%$). 

\begin{figure*}[!htb]
\centering
\subfloat[][]{\includegraphics[scale =0.8]{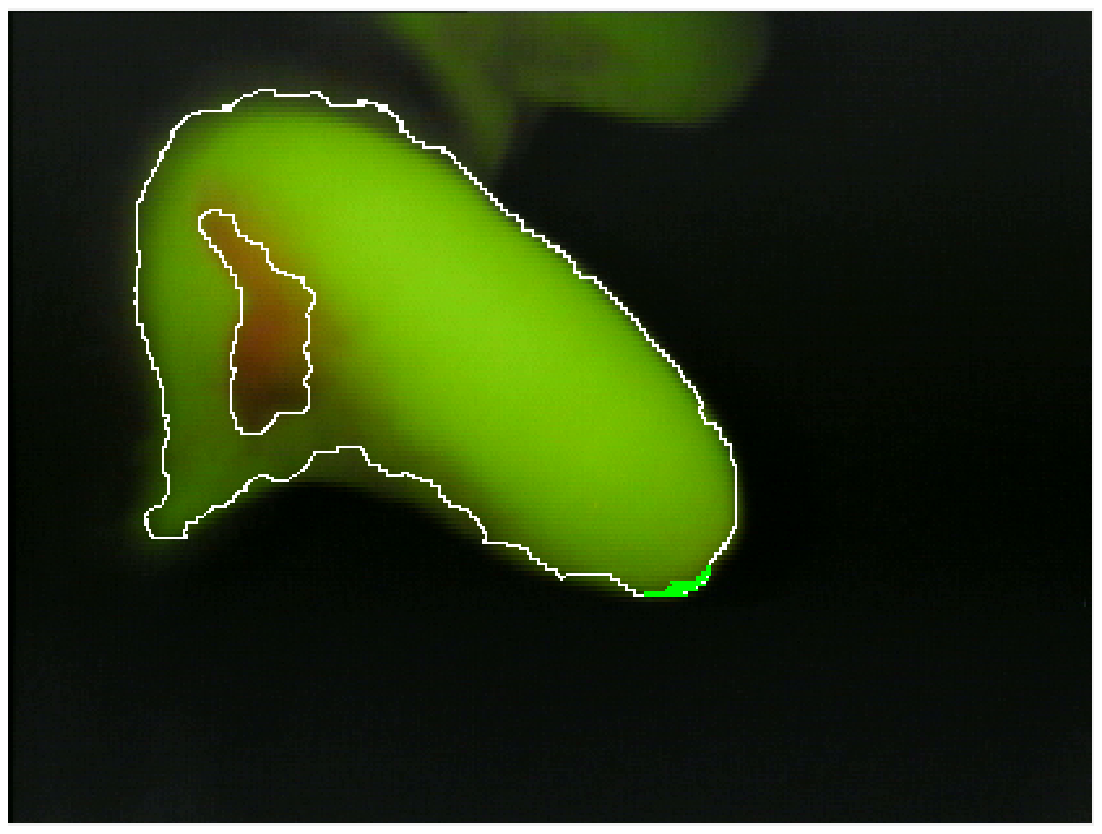}\label{DjadaL28I7BR}}
\subfloat[][]{\includegraphics[scale =0.8]{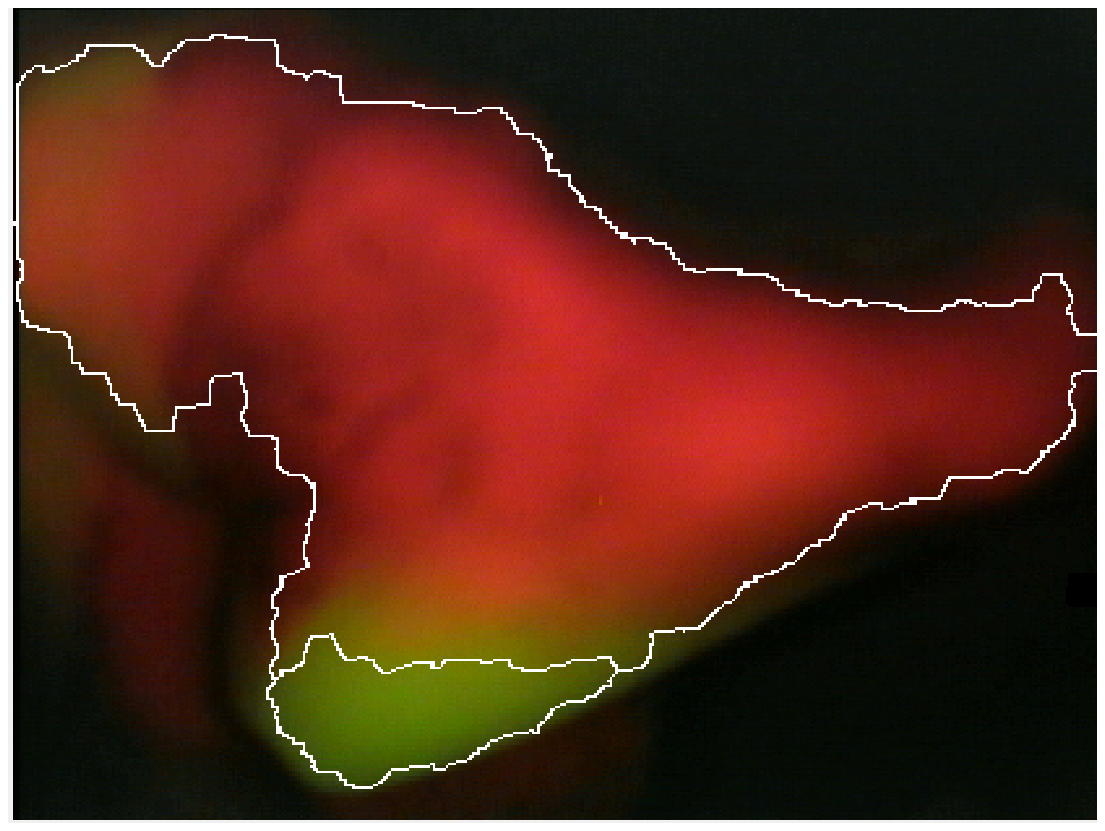}\label{Dh2b029L14I6BR}}\\
\subfloat[][]{\includegraphics[scale =0.8]{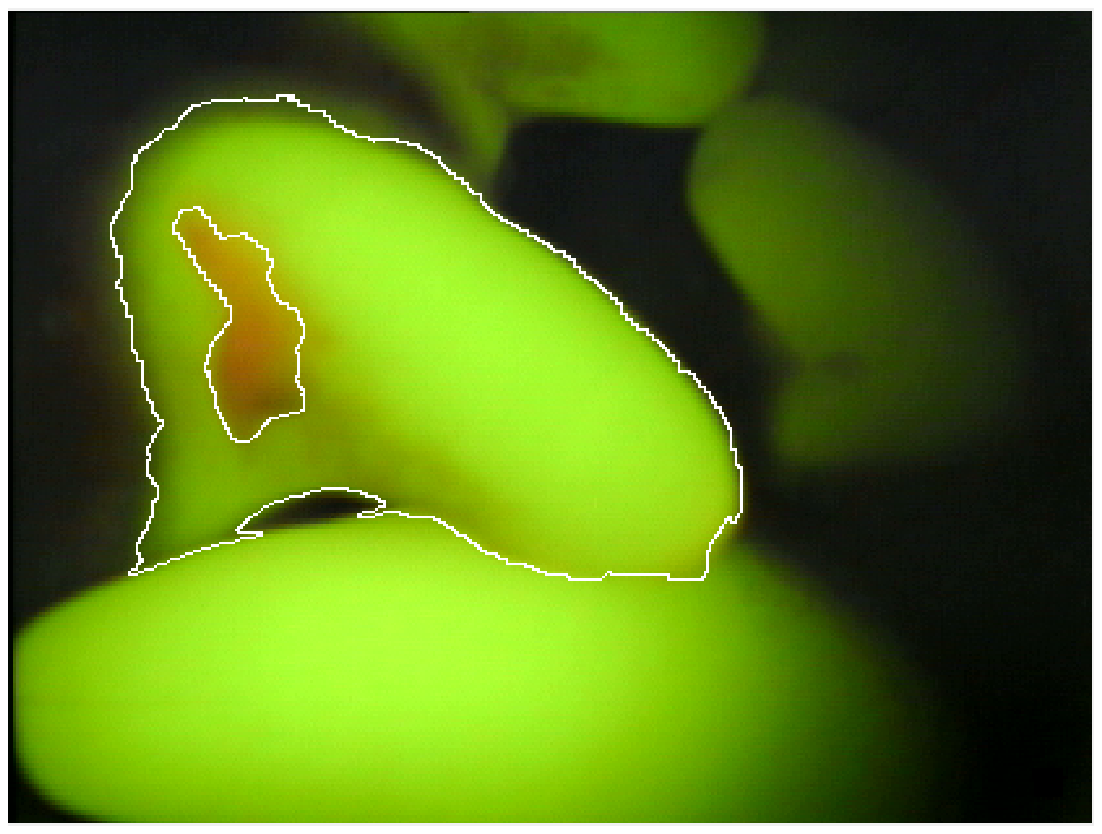}\label{DjadaL28I7BR-3}}
\subfloat[][]{\includegraphics[scale =0.8]{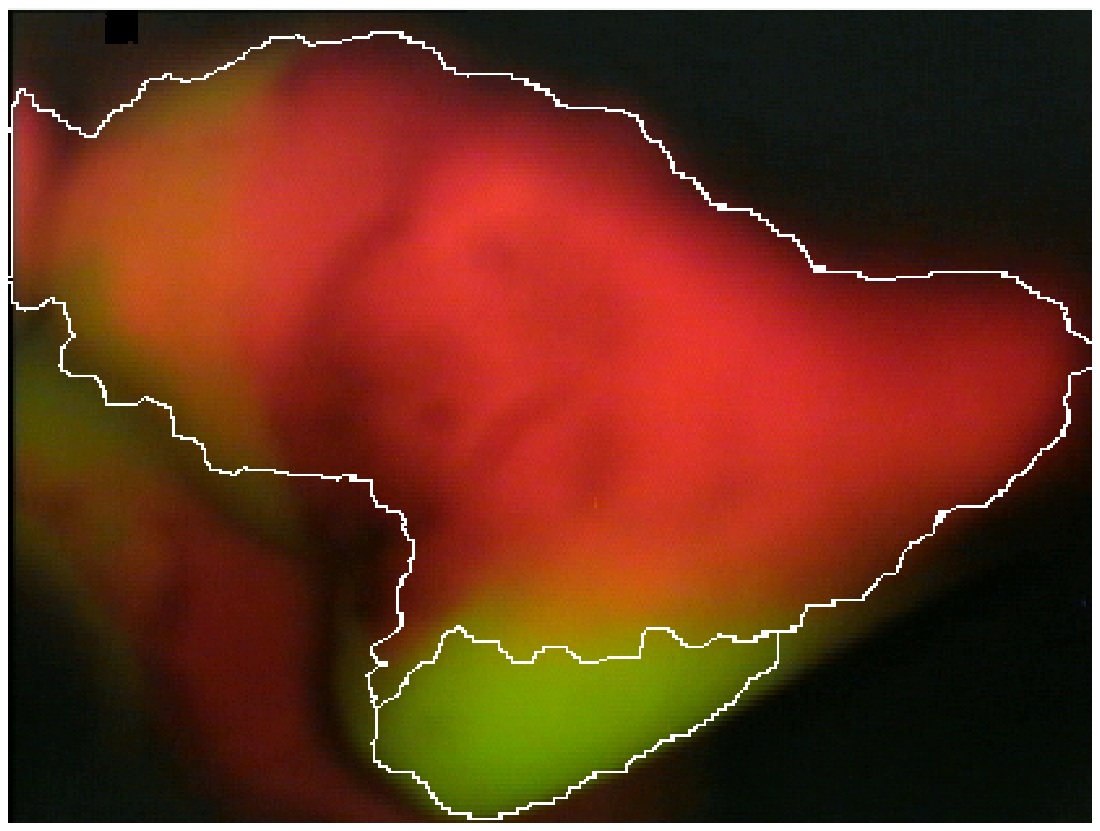}\label{Dh2b029L14I6BR-3}}\\
\caption{The quantification maps produced by the proposed automatic quantification algorithm for QLF images (pre-expert tweaking). Two teeth from the same subject are scanned over a period of 6-months to quantify plaque over time. (a) shows a premolar canine tooth at month 0, (c) shows the same tooth after 6 months. (b), and (d) show a molar teeth of a canine imaged at the start of the study (month 0) and month 6 respectively.}
\label{failedimages}
\end{figure*}

The biofilm quantification index computed during the first phase of computer-aided biofilm quantification closely follows manual scoring results. The algorithm has been evaluated on 470 images. The results are highly consistent with the manual gradings. Fig. \ref{failedimages} shows an initial-guess quantification on a set of dental images for two teeth taken over a period of 6 months. 
To evaluate the agreement between the quantification performed manually by human graders and the initial automated segmentation, we employed \emph{Cohen's kappa coefficient}, which measures the agreement between two raters (ground truth and the proposed method) who are classifying $N$ terms into $C$ mutually exclusive classes. The kappa coefficient $\kappa$ is defined as
\begin{equation}
\kappa  = \frac{{\Pr \left( a \right) - \Pr \left( e \right)}}{{1 - \Pr \left( e \right)}}
\end{equation}
where $\Pr \left( a \right)$ is the relative observed agreement among raters, and $\Pr \left( e \right)$ is random agreement among raters. If there is a perfect agreement among raters then $\kappa=1$, and if there is no agreement except by chance (defined by $\Pr \left( e \right))$ then $\kappa=0$. We applied $\kappa$ statistics to our data set ($N=470$, $C=8$), and obtained  $\Pr \left( a \right)=0.6827$, $\Pr \left( e \right)=0.1389$, and $\kappa=0.6315$. According to \cite{Gwet2010}, values $<0$ indicate complete disagreement, $0$ to $0.20$ slight agreement, $0.21$ to $0.40$ fair agreement, $0.41$ to $0.60$ moderate agreement, $0.61$ to $0.80$ substantial agreement, and $0.81$ to $1.0$ perfect agreement. Table \ref{table:ConfusionMatrix} shows the confusion matrix for actual (\emph{ground truth}) and predicted (\emph{automatic quantification algorithm}) classes. Furthermore, we created 8 one-versus-all classifications by categorizing images as either belonging to their true class or not. For an $n$-way classifier, a one-versus-all classifier can be constructed automatically from an $n\times n$ confusion matrix by treating the image to be classified as belonging to the class if the class is the result of classifying it correctly.
\begin{table}
\begin{center}
\begin{tabular}{cc|cccccccc}
\hline
&&\multicolumn{8}{c}{\textbf{Predicted Class}}\\
&&1&2&3&4&5&6&7&8\\
\hline
\multirow{8}{*}{\begin{sideways}\textbf{Actual Class} \end{sideways}}
&1&64&9&0&0&0&0&0&0\\
&2&8&57&7&0&0&0&0&0\\
&3&0&     6&     51&     4&     0&     0&     0&     0\\
&4&0&     0&     7&     56&     9&     0&     0&     0\\
&5&0&     0&     2&     5&     48&     9&     0&     0\\
&6&0&     0&     0&     0&     10&     52&     0&     0\\
&7&0&     0&     0&     0&     0&     9&     49&     5\\
&8&0&     0&     0&     0&     0&     0&     0&     0\\
\hline
\end{tabular}
\end{center}
\caption{The confusion matrix for the canine teeth with 8 actual classes provided by the human grader. In order to have a comparative study, automatic biofilm quantification results from the algorithm are mapped as 8 \emph{predicted} classes.}
\label{table:ConfusionMatrix}
\end{table}
For most cases the proposed algorithm provides a fairly accurate assessment of the biofilm; however, there could be instances where the automatic quantification fails to perform optimally and therefore needs the additional integrated manual supervision provided by the second stage of our method. We found two image types that produce consistent misclassifications: the first includes images with multiple teeth that are fluorescing with similar intensity, thus making it difficult to define a clear boundary between them; the second includes images containing reddish artifacts appearing at the boundary between tooth and gum. Owing to relatively simple texture in QLF dental scans, the number of superpixels does not appear to have much impact on the performance of the method over a wide range. The adequate number of superpixels depends on the ratio of size of the image to size of captured tooth of interest inside the image as well as the amount of fluorescent from neighboring teeth.

\begin{figure*}[htb]
\centering
\includegraphics[scale =1.0]{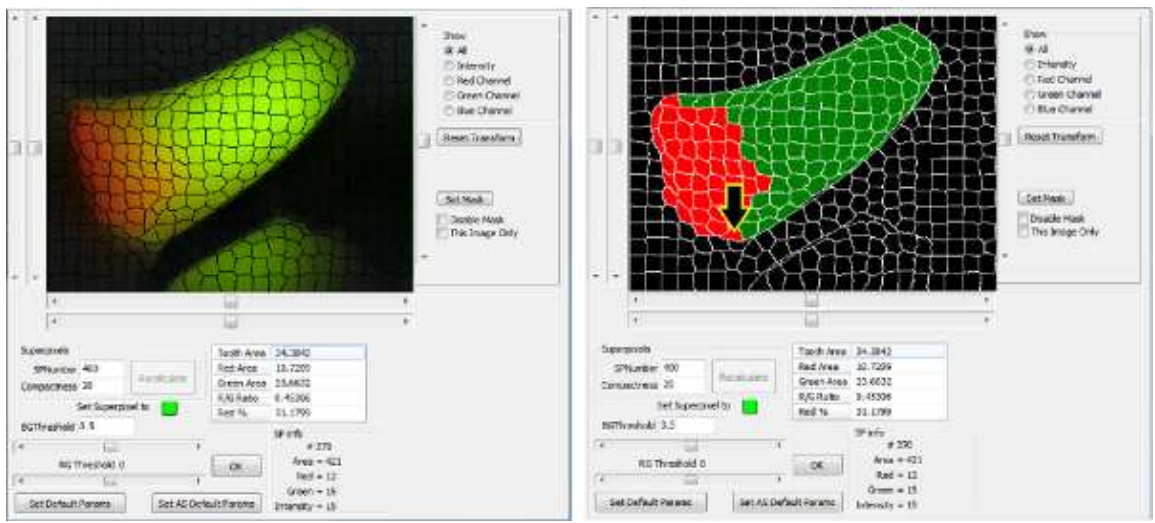}
\caption{Screenshot of the software implementing the algorithm, showing the quantified tooth before manual adjustment. The initial automatic quantification (\emph{left}), if deemed suboptimal (red arrow), can be tweaked by the expert by a simple \emph{click-and-flip} approach to a different class (\emph{biofilm}, \emph{tooth subtrata}, \emph{background})(\emph{right}).}
\label{fig:screenshot2}
\end{figure*}

\begin{figure*}[htb]
\centering
\includegraphics[scale =1.1]{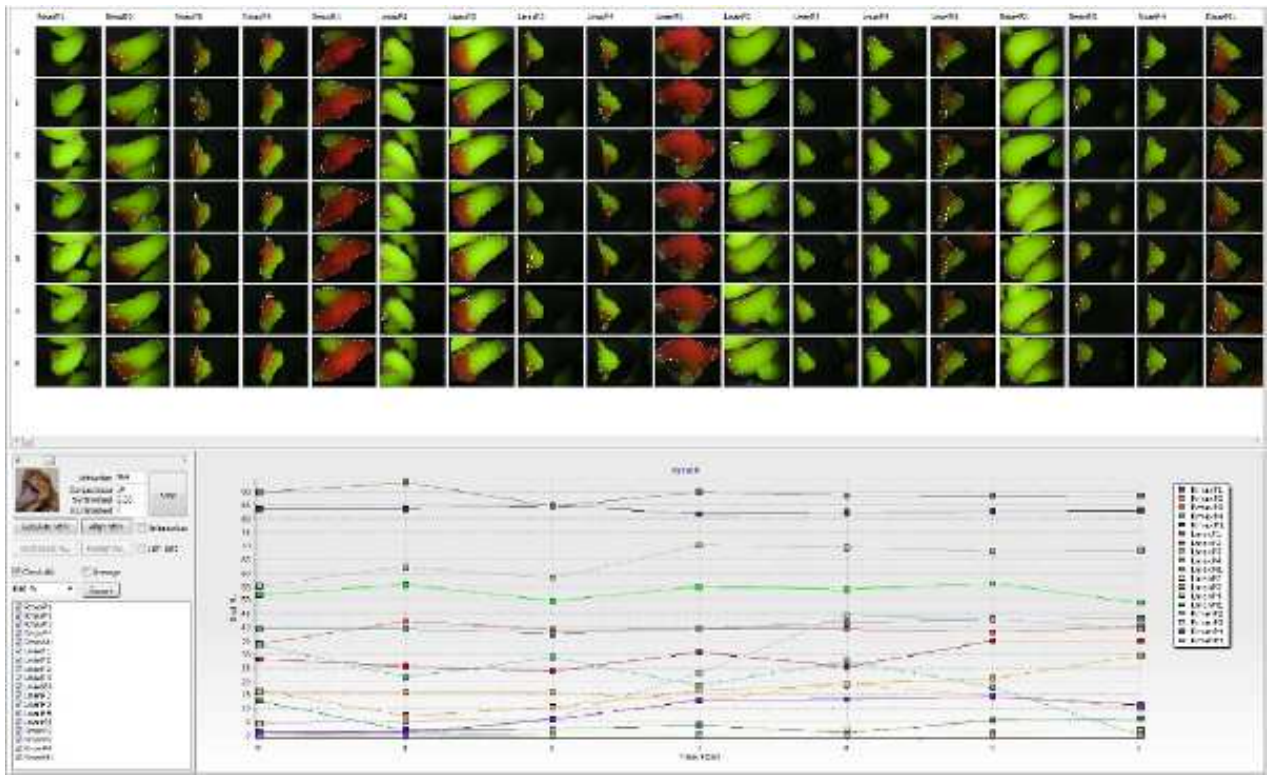}
\caption{The \emph{batch-mode} of the proposed method showing the entire jaw-set being processed at once. The expert correction if needed due to non-uniformities in the acquisition procedure needs to adjusted once only on a single representative image for a given capture settings.}
\label{fig:screenShotscreenShot}
\end{figure*}

The images with misclassified superpixels constitute about $1\%$ of our testing set. We observed that it takes a human grader interacting with initial unsupervised guesses 5 seconds per image on average to fix misclassifications. Screen shots of the software are shown in Fig. \ref{fig:screenshot2} (\emph{left} fully automatic first stage , and \emph{right} second stage optional correction). The software can be downloaded from here (\url{https://www.nitrc.org/projects/biofilmquant}). The current version of the program has the capability to process an entire jaw set in batch mode along with the longitudinal analysis (Fig. \ref{fig:screenShotscreenShot}). The longitudinal analysis of the proposed technique provide a tool to help understand the progression of biofilm as well as the monitoring of the effectiveness of clinical procedure over time. The software also has the capabilities to output the label image, the total biofilm coverage, and the mean biofilm coverage over time.

Our tests demonstrated that the utilized \emph{click-switch} approach of superpixel re-classification is much more intuitive and accurate than drawing boundaries around tooth and plaque. The Fitt's Law providing an empirical model of human muscle movement (primarily used in human-computer interaction) explains the speed-accuracy trade-off characteristics: the faster we move, the less precise our movements are; the stricter the constraints are, the slower we move. According to Fitt's Law, the \emph{click-switch} approach is expected to be faster and more accurate than drawing boundaries by order of logarithm.

\section{Conclusion}
This paper presented a semi-automated method for biofilm quantification in canine teeth that is independent of grader and instrument bias. The method takes a leap from traditional techniques of manual demarcation of tooth and plaque boundaries; at the same time it avoids simplistic automation approaches that allow pathologists only to completely accept or reject the quantification these methods produce. The core of the approach is a Gauss Markov random-field statistical model for QLF images. The results demonstrate that biofilm quantification using the statistical model is robust, reliable, and reproducible. The segmentation of QLF images into superpixels makes the classification and the subsequent quantification process accurate and fast. The method provides an initial quantification guess that is found to match human expert judgment in $99\%$ of the tested images. The misclassified images can be manually corrected by an expert using a single-click switching of superpixels among three classes (background, tooth substratum, and dental plaque). The approach could be extended further to other quantification methods such as dental fluorosis by incorporating more features. Dental fluorosis is caused by excessive exposure to high concentration of fluoride. Fluorosis often appears as tiny white streaks or specks in the enamel of the tooth, and in extreme forms tooth appearance is marred by discoloration and brown markings. Current method of quantification is 6-class rater-based H.T. Dean's fluorosis index \cite{american1993fluoridation} proposed in 1942. More accurate quantification methods such as proposed here in this paper can help early diagnosis and better treatment. Furthermore, using expert corrections as a feedback with machine-learning methods for adjusting the cut-off threshold can help refine the initial segmentation.

\ifCLASSOPTIONcaptionsoff
  \newpage
\fi


%

\bibliographystyle{IEEEtran}
\bibliography{refsDental2}

\end{document}